# Review: Metaheuristic Search-Based Fuzzy Clustering Algorithms

[1]Waleed Alomoush, [2]Ayat Alrosan

Imam Abdulrahman Bin Faisal University

**Abstract** *Fuzzy clustering is a famous unsupervised learning method used to collecting similar data elements within cluster according to some similarity measurement. But, clustering algorithms suffer from some drawbacks. Among the main weakness including, selecting the initial cluster centres and the appropriate clusters number is normally unknown. These weaknesses are considered the most challenging tasks in clustering algorithms. This paper introduces a comprehensive review of metahueristic search to solve fuzzy clustering algorithms problems.*

**Keywords** *Fuzzy clustering, Metaheuristic search, Local Search.*

## 1. Introduction

Fuzzy clustering is a partitional clustering technique that is based on the objects of a fuzzy soft set theory. The fuzzy soft set theory holds the notion that for a given universe of discourse, every object in the universe belongs to a varying degree to all sets defined in the universe [59, 70, 71]. In fuzzy clustering, the universe of discourse is all the objects in the given dataset and the sets defined on the universe are the clusters. Objects are not classified as belonging to one and only one cluster, but instead, they all possess a degree of membership with each of the clusters. Clustering is a unsupervised learning approach that is capable of partitioning identical data objects (patterns) based on some level of similarity, which increases the similarity of objects within a group and decreases the similarity among objects between various groups[13, 15, 37, 38]. Fuzzy clustering has several advantages that make it a preferable clustering algorithm. These advantages are fuzzy natural of FCM provides the algorithm with more information on the given dataset, simple and straightforward programming implementation, suitable for very large datasets since its time complexity is O(n), produces very good results in some conditions (i.e., hyper spherically shaped well-separated clusters) and robust and is proved to converge to local optimal solution. However, fuzzy clustering has some weaknesses such as the number of clusters in the given dataset should be known a priori, the sensitivity to the cluster centres initialization phase and sensitive to noise and outliers. This paper is organized as follows: section two provides an review of cluster centres Initialization Sensitivity problem; section three provides an review of determining the number of clusters problem and the final section consist of discussion part

## 2. Cluster Centres Initialization Sensitivity - Local Optima Problem

The selecting initial cluster centre value is considered one of the most challenging tasks in partitional clustering algorithms. Since incorrect selection of initial cluster centres values will make the searching process towards an optimal solution that stays in local optima, and therefore produce undesirable clustering result [14, 33]. The main cause of this problem lies in the way that the clustering algorithm works as its run in a manner similar to the hill climbing algorithm [40]. The hill climbing algorithm is a local search-based algorithm that moves in one direction without performing a wider scan of the search space to minimize (or maximize) the objective function. This behaviour prevents the algorithm to explore other regions in the search space which might have a better, or even the desired solution. Figure .1 is a graphical demonstration of the local optimal problem.

A common and simple approach to alleviate this problem is to re-run the algorithm several times with several cluster initializations. However, this method is inapplicable in many cases such as a large dataset, or complex dataset (i.e., dataset with multiple optima) [35]. Hence, the use of optimization algorithms in solving clustering problem is preferable.

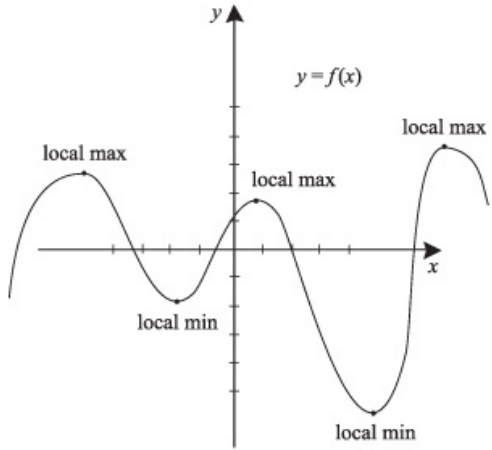

Figure 1. Local optima problem.

In the spirit of the main cause of this problem, the local search behaviour, several global based or improved local-based search algorithms were proposed in the last many decades to address this problem [12, 32, 40, 51, 58, 62]. Indeed, these algorithms include local search-based metaheuristic algorithms such as SA, TS or population-based metaheuristic algorithms such as EAs (including Genetic Programming, Evolutionary Programming (EP), Evolutionary Strategies (ES), GAs and DE), HS or swarm intelligence algorithms such as PSO, Bees algorithm, Artificial Bee Colony (ABC) and ACO. More details on metaheuristic approaches are given in Appendix A. The main goals of population-based search algorithms are their capabilities to deal with local optima and explore huge solution spaces effectively by maintaining, recombining and comparing many candidate solutions at the same time. While the local search-based nature inspired algorithms have some advantages over classical local search algorithms in coping with the local optima problem, their results are somehow weak compared with population-based metaheuristic algorithms [20]. The following is an overview of these algorithms proposed to solve clustering problem where the clusters number in known or set up a priori.

## 2.1 Local Search-based Metaheuristic Clustering Algorithms

The local search-based metaheuristic algorithms were used to solve such clustering problem as in the following:

In [62], the authors discussed the deal of the clustering problem always solved by the K-means technique. The weakness is obtained local minimum solutions, which are always what the K-means technique performs. The SA method for address optimization problems is proposed for solving the clustering weakness. The factors of the technique are discussed in full detail and it is shown that the technique converges to solution of the clustering weakness. Al-Sultan [1] proposed a clustering technique with a TS heuristic and showed that the proposed technique excel both the K-means technique and the SA-based clustering technique [1]. The author also proposed the same algorithm but for the fuzzy clustering problem [2]. An improved TS-based clustering algorithm was proposed by Sung and Jin [65]. Their proposed technique is a combination of the TS heuristic with two complementary functional procedures, called packing and releasing procedures. The technique was numerically tested for its electiveness in comparison with reference works including the TS algorithm, the K-means technique and the SA technique. These local search metaheuristics, TS and SA, just refine one candidate solution and are weak in coping with local optima. These technique are particularly factors sensitive and tuning is highly problem-dependent [57].

## 2.2 Population-based Metaheuristic Clustering Algorithms for Fuzzy Clustering

The population-based metaheuristic algorithms have been intensively used to solve such problem for fuzzy clustering. In the following sections, a description of such algorithms based on fuzzy clustering, is presented. Actually, the EAs for fuzzy clustering are somehow normally built upon adaptations of current methods developed for hard clustering. These algorithms can be broadly categorized into two approaches [35]. The first approach which is the most representative one is composed of two steps: first finding the appropriate cluster centres via EA, and secondly, the FCM algorithm performs clustering with initial cluster centre values obtained from the first step. The second approach of evolutionary fuzzy clustering techniques is composed of using the EA as a clustering algorithm by itself. Some of these algorithms use FCM or its variants as a local search engine to support the performance of their algorithm by speeding up their convergence and fine-tuning the evolved solutions that were explored by the evolutionary search. This categorization is also applicable to metaheuristic-based hard clustering algorithms. Some of these algorithms are discussed below:

**GA-based Fuzzy Clustering:**

Hall, Bezdek [29], Hall and Ozyurt [31] and Hall, Ozyurt [32] showed that their proposed genetic-guided clustering algorithm can be a good clustering tool. GA was used to find the near optimal cluster center values where, FCM/HCM is used to perform the clustering. A real encoding scheme is used to encode cluster centres in GA's population. A reformulated version of standard c-means' objective function that exclude partitioning matrix U is used as a fitness function to control the GA's evolving process. Their proposed method showed a promising result compared with random initialization of FCM/HCM. They also demonstrated that their GA method does not appear to have the capability to produce focused small improvement to the cluster centres to support a final clustering as in FCM/HCM. However, the GA guided clustering technique is very sensitive to the initial random solution that is used to generate an initial population. Furthermore, not enough experiments were conducted that can show the ability of the proposed algorithm to avoid a premature convergence problem. Liu and Xie [45] showed that their proposed genetics-based fuzzy clustering approach have much higher probabilities for reaching the global optimal solutions than the traditional technique. GA is used as a fuzzy clustering algorithm in this study and named as GFCM, while the evolving process was based on using FCM's objective $(i.e., J_m(U,V))$ function as fitness function. All standard GA's operators were used with a binary encoding scheme for cluster centres representation in each chromosome. In their experiments, the authors showed that GFCM has better results than FCM; however there is a limitation encountered regarding the local optima problem especially in the small populations. In some cases, GFCM gets stuck in local optimal solution. Van Le [67] proposed two approaches for the fuzzy clustering problem. These approaches are based on GAs and EP respectively. He concluded that these methods succeed in many cases where the FCM algorithm failed, and that the EP method appears to produce the best results.

Klawonn and Keller [43] investigated an EA for clustering different types of cluster shapes such as shell and solid clusters (rectangle and cubes). They reported that such algorithm cannot produce promising results for shell shape but can be helpful for the other shapes. No details of the evolutionary clustering algorithm were given however.

Egan, Krishnamoorthy [24] proposed a GA-based fuzzy clustering algorithm for noisy data. Their proposed algorithm is different from other GA-based clustering algorithms since it is directed to cluster different levels of noise data. An extra cluster that represents noise data, called noise cluster, is added to each chromosome. The authors reported that the binary representation outperformed a real-valued representation in this domain, while Hall, Ozyurt [32] reported that this conclusion is not documented. Their algorithm showed a good result in case of less noise data compared with the so called robust FCM [22] while in case of noise data, the algorithm performed poorly [32].

Maulik and Saha [51] proposed a modified DE-based fuzzy clustering technique which they named (MoDEFC) [51]. The mutation process of standard DE was modified in this algorithm through the use of the concepts of local and global best vector of PSO algorithm. These concepts were introduced during the mutation process to push the trial vector quickly towards the global optima. Here, the global best (GBest) vector among all DE generations and local best (LBest) vector among current generation are introduced to replace the random selection of these two vectors in the mutation process. Their proposed algorithm was used as an image segmentation algorithm. In addition, several synthetic and real life datasets as well as some benchmark functions were used to show wide applicability of the authors' proposed algorithm.

**ACO-based Fuzzy Clustering:**
The ACO-based clustering algorithm has been proposed by [39] and [40]. Their proposed algorithm uses the concepts of ACO to find near optimal partitions of the data. The proposed method is similar to what has been done in [32]; it compress two phase: first, the algorithm, ants optimization mechanism, is used to find near optimal cluster centrees values, while in the second phase, the algorithm uses these values as the initial canters of cluster for the FCM algorithm to perform the clustering process. As in [32], they used the reformulated FCM's objective function as a fitness function to control the optimization process of ACO. A set of benchmark data was used in their experiments such as iris, glass, cancer, etc. Promising results were shown compared with FCM algorithm. However, Supratid and Kim [66] showed some limitations of this algorithm regarding Pheromone, t value that is updated without any feedback from the procedure itself. Furthermore, the disconnection between ACS and FCM after ACS find the near optimal cluster centrees may lead to some weaknesses. Supratid and Kim [66] proposed a modified ACO-based fuzzy clustering algorithm which they named (MFAC). Their approach combines ACO, FCM and GA all together. The main idea of the combined this algorithms hinges on the frequently improvement of cluster outcomes. The authors reported that using a GA to optimize the initial clusters for FCM is still unsatisfactory. Since a GA increases just the current better solution, it will thus easily get trapped in local minima. The MFAC algorithm starts by using ACS for the cluster centrees refinement process during which the algorithm calls another procedure named GFCM. This algorithm is a combination between GA and FCM similar to what was proposed in [32]. Therefore, the main difference among the integration implemented in this work and the one done in [40] is that in this algorithm the integration of ACS with FCM occurs many times (through calling GFCM) during the clustering process whilst the integration of ACS with FCM optimizes the FCM initial clusters only once. Depending on the fact that the performance measurement relates to the percentages of accuracy of image retrieval, the researchers' experimental results showed that their proposed method yields the best outcomes between others in creating fuzzy histogram in image retrieval application.

**Bee-based Fuzzy Clustering:**
Pham, Otri [58] proposed a Bees-based fuzzy clustering technique. Their proposed clustering approach exploits the search ability of the Bees Algorithm to control the local optimum weakness of the FCM technique. particularly, the mission is to search for appropriate cluster centres, which are represented as a real number in each bee, such that the FCM's objective function is minimized. They compared the performance of their proposed algorithm with FCM and GA-based clustering algorithm on some numerical benchmark data. They showed that Bees-based clustering algorithm outperforms the others. Karaboga and Ozturk [42]proposed an ABC algorithm as a fuzzy clustering algorithm. They compared the performance of their proposed algorithm with FCM algorithm on some numerical benchmark data. Their results proved that ABC-based clustering algorithm outperforms FCM algorithm. [7] proposed artificial bee colony based fuzzy clustering algorithms for MRI image segmentation

**PSO-based Fuzzy Clustering**
In 2003, Xiao et al. used a new method based on the PSO and the Self Organizing Map (SOM) for clustering the gene expression data [69]. In their proposed method, the rate of convergence is improved by adding a conscience factor to the SOM algorithm. The proposed hybrid SOM/PSO approach uses PSO to develop the weights for SOM. The weights are trained by SOM in the first phase. In the second phase they are improved by PSO. The authors could achieve promising outcomes by applying the hybrid SOM-PSO approach over the gene expression data of Yeast and Rat Hepatocytes. Cui, Potok [19] and Cui and Potok [18] proposed a PSO based hybrid algorithm for classifying the text documents. The hybridization step is basically a two-step clustering approach where in the first step the PSO performs clustering until the maximum number of iterations is exceeded. Then in the second step the k-means algorithm is initialized with cluster centers obtained from the first step and then performs the final clustering procedure. This combination of the two steps is meant to improve the PSO performance and speed up its convergence especially in the case of large datasets. The authors applied the K-means, PSO and a hybrid PSO clustering technique on four different text document datasets and their outcomes clarify that the hybrid PSO technique can generate high compact clustering outcomes over a short span of time than the K-means algorithm.

**HS-based Fuzzy Clustering:**
Recently, the HS algorithm has been used for clustering problems as reported by Ingram and Zhang [36]. many researchers combined FCM algorithms with HS (FCM/HCM) in one framework such as Ayvaz [8] who combined HS with FCM to estimate the region structure and region transmissivities for a heterogeneous aquifer, Forsati, Mahdavi [26] and Mahdavi, Chehreghani [47] both developed the performance of HS for web documents clustering using the integration of k-means technique as a local search formed. This is done use calling k-means technique a few times then k-means obtained the clustering and the returned vector is added to HM if it has a best objective value than those in HM. In the same year; Fuzzy classification of the Fisher Iris dataset was fulfillment by Wang, Gao [68]. Initial classification was obtained by the (FCM) algorithm then the optimization of the fuzzy membership functions was calculated by a new hybrid HS-Clonal Selection Algorithm (CSA) approach. In 2008 also; Malaki, Pourbaghery [49] improved two hybrid IHS-FCM clustering approach, which have been tested on a 58,000 elements NASA radiator dataset. In their proposed approach, the authors used the improved version of HS that was proposed by Mahdavi, Fesanghary [48] (IHS), in addition; they integrated it with FCM algorithm to develop its accomplishment. FCM in fact is integrated in two methods: first, it is integrated into IHS as a local search formed to maximize the convergence speed similar to what was employed by Mahdavi, Chehreghani [47]. This way is named (FHSClust). In the second method, FCM is used as a further final clustering phase to improve the partitioning outcomes, where it is created by the best solution vector improvised in FHS Clust.

Alia, Mandava [4] proposed a new approach to tackle the well-known fuzzy c-means (FCM) initialization problem. This approach uses a metaheuristic search method called Harmony Search (HS) algorithm to produce near-optimal initial cluster centers for the FCM algorithm. To demonstrate the effective this approach by a MRI segmentation problem. The experiments indicate encouraging results in producing stable clustering for the given problem as compared to using an FCM with randomly initialized cluster centers.

**FA-based Fuzzy Clustering:**
Senthilnath, Omkar [63] employed FA for the purpose of clustering data objects into groups as per the values of their attributes. The performance of FA for clustering was then compared with those of other nature-inspired algorithms such as artificial bees colony (ABC) and particle swarm intelligence (PSO). The comparison also looked into the performance of nine other algorithms and approaches as utilized in the established literature on the 13 test data sets specified in [42]. The main performance criteria exploited for the purpose of comparison was the classification error percentage (CEP). CEP can be defined as the ratio of the number of misclassified samples in the test data set to the total number of samples in the same set. After extensive testing the researchers surmised that FA was the most effective method for clustering. [5, 6] proposed the fuzzy clustering algorithm using hybrid the firefly algorithm FA with fuzzy c-means algorithm FCM which is called FFCM to produce a new segmentation method for real and simulated MRI brain images

## 3 Determining the Number of Clusters

Partitional clustering techniques suppose that the clusters number in unlabelled data is known and specified a priori by the user. Generally, this number is unknown, especially for high d-dimensional data, where visual investigation is inapplicable [3, 12, 22]. Indeed, selecting the convenient clusters number is a non- easy task and can lead to undesirable outcomes if inaccurate executed. Thus, several research were proposed to solve this matter

**GA-based Dynamic Clustering:**
Maulik and Bandyopadhyay proposed an algorithm named FVGA for automatic determination of an appropriate number of clusters with the corresponding fuzzy clustering results [50]. In their proposed algorithm FVGA, they used GA associated by XB cluster validity index as a fitness function to determine which chromosome will be evolved. This algorithm uses a real scheme to encode each candidate partition such that each genotype (chromosome) contains candidate cluster centres. To enable FVGA to determine the appropriate number of cluster centres, the authors used a variable length concept for each chromosome where each chromosome in the population can encode a different number of cluster centres. Thus, each chromosome has a number of cluster centres ranged over predefined values [min, max]. Furthermore, the single point crossover operator is used and modified in this study to guarantee that the offspring has at least two cluster centres (the minimum value of the range). The other GA operator, mutation operator, is subjected to each gene (cluster centre) in the chromosome when it is within the mutation probability mm. Finally, FVGA is hybridized with one step recalculating of clustering centres using the FCM's cluster centre Eq. 5.10 as a mechanism to fine-tune each candidate partition. The hard version of FVGA can be found in [9, 12] under genetic clustering for unknown K (GCUK-clustering). The GCUK algorithm is the same as FVGA except that it used the DB index [23] as an objective function. In the same context, Pakhira, Bandyopadhyay [55] slightly modified the same algorithm by fixing the length of each chromosome through making its length equal to a predefined expected maximum number of clusters in the given dataset, while at the same time maintaining the idea of variable length. This is presented by incorporating the concept of empty gene in the chromosome that is represented by using the expression of 'don't care'. The other GA operators are the same as original FVGA except that the selection process changed to tournament of size 2. The main objective of this research is to evaluate FVGA with a variety of cluster validity indices (PC, PE, XB and PBMF). In addition, another comparison between FVGA [50], FCM and GGAFCM [32] was conducted based on measuring the performance of these algorithms in terms of validity indices. They demonstrated that the PBMF cluster validity index is better able to indicate the appropriate clusters number in a dataset irrespective of the source clustering algorithms used. Furthermore, they found that the efficiency of FVGA approach is to be comparable to, always better than the classical FCM technique and GGAFCM approach. However, FVGA in both researches[56] and[50] has some limitations as follows:

The standard GA operators such as single point crossover and mutation may lead to an invalid solution in terms of region density representation (e.g., representing a region with low data density or the same number of cluster centre but with repeated centre values) that may affect the speed of convergence or the quality of the solution. This problem is highlighted by [35]and named "context-insensitivity". This term "context-insensitivity" means as in [25] that "the schemata defined on the genes of the simple chromosome do not convey useful information that could be exploited by the implicit sampling process carried out by a clustering genetic algorithm". In general, standard GA operators are usually not clustering oriented and should be modified to be more suitable for these types of problems [25, 35].

Saha and Bandyopadhyay proposed a new fuzzy dynamic clustering algorithm known as Fuzzy-VGAPS [60, 61]. This new algorithm is, in fact, based on the same basis of FVGA algorithm [50, 55] except for some modifications made to the objective function used in calculating the quality of the evolved chromosomes as well as some modifications to the GA operators. For the objective function, they used a new point symmetry-based index called fuzzy Sym-index which is the fuzzy version of Sym-index proposed by the same authors in [11, 60]. This index is actually a modified version of the original PS-index proposed by [16, 17]. Furthermore, the membership grade for each data point to a particular cluster centre is calculated based on a conditional parameter called Q to use either a point symmetry distance or Euclidean distance. The probabilities of mutation and crossover operators are selected adaptively as in [64]. The hard version of this algorithm (Fuzzy-VGAPS) can be found in [11] under (VGAPS) name. VGAPS is similar to Fuzzy-VGAPS except that VGAPS used the hard version of the objective function named Sym-index. The outcomes of the Fuzzy-VGAPS are compared with those performed by other technique consist of both crisp and fuzzy techniques on four real-life datasets and four artificial consist of magnetic resonance image (MRI) brain with multiple sclerosis lesions. Despite the promising results compared with other algorithms, Fuzzy-VGAPS still has some limitations listed as follows:

- Fuzzy-VGAPS will probably fail if the clusters do not have (symmetry) property since the similarity measure used in this algorithm is based on this feature [10].
- The fixed near value (which is the unique nearest neighbour of symmetric point used in this algorithm) may lead to many drawbacks.
- Fuzzy-VGAPS is still very time consuming [41].

- Fuzzy-VGAPS has the same limitations of GA operators as mentioned for FVGA algorithm.

**DE-based Dynamic Clustering:**
Das and Konar [21] proposed a new algorithm for fuzzy dynamic clustering with application to the image segmentation problem. Their proposed algorithm named (AFDE) is based on DE optimization algorithm. AFDE is applied to six different types of images including MRI brain image, natural image, and satellite image. PS index [16, 17] was used as an objective functions to evaluate the solutions generated during the evolution process. In this study, the authors modify the standard DE algorithm in a way to be more suitable for the dynamic clustering problem and at the same time avoid the traditional EAs problem known as stagnation and/or premature convergence. For the later, the authors first modified the mutation constant factor 'F' and made it randomly ranged between 0.5 and 1. Secondly, instead of being fix during the evolving process, the crossover rate Cr value has been modified to be variable, therefore, started from rate value 1 and linearly decreased till it reached the minimum acceptable value which is 0.5 over the time. For the former, the authors modified the chromosome representation in a way that can make the AFDE dynamically determine the appropriate number of clusters that the dataset may have. Each chromosome encode (using real encoding as original DE) a candidate number of cluster centres where the length of it is determined previously by user, which represents the expected maximum number of cluster centrees that the test dataset may possess. The hard version of this algorithm (AFDE) can be found in [20] named as ACDE. ACDE is the same as AFDE except that ACDE use different objective function named the CS index [17] and also they used the DB index [23] as an alternative option. A new variant of ACDE is proposed by Gong, Cai [27] named ACDEPS, where the authors used the same framework as ACDE except for the objective function. They replaced the CS index [17] with slight modification on Symindex [11] in calculating the fitness function of each evolved chromosome.
Limitations of AFDE:
The objective function (PS index) that is used in AFDE has some limitations and may lead to undesirable results as reported by [10]. These limitations are summarized as follows:

- The computational load required by PS index is $O(c.n^2)$ that becomes very expensive as the number of cluster centrees c and data points n increase [10, 20] (e.g., in case of 512×512 image, the number of pixels = 262144).
- The PS validity index may produce unwished for clustering results. This is because the PS index may fail to determine the proper cluster centrees for data point when the clusters themselves are symmetrical with respect to some intermediate point [10].

**ACO-based Dynamic Clustering:**
A dynamic fuzzy clustering algorithm based on ACO was proposed by Gu and Hall [28]. They replace the Euclidean distance that is normally used as similarity measurement in FCM with kernel-induced distance metric. This modification is used to overcome the FCM shortcoming when the cluster shapes are not hyper-spherical. A reformulated kernel induced distance-based XB index is proposed in this study for measuring the improvement of the optimization process (i.e., fitness function). The reformulated version is based merely on calculating cluster centres and excludes calculating membership matrix U. The coordinate of ants to move centres of cluster in feature space to explore for optimal cluster centres. Promising results were obtained by applying the ACO-based clustering algorithm over three datasets, the Iris data, an artificial five classes dataset and a one feature MRI image with three classes (i.e., CSF, WM, and GM). The same technique was also proposed by Hall and Kanade [30], a swarm based fuzzy clustering technique by the XB partition validity metric, which find the clusters number wonderful for many datasets. They keep the Euclidean distance as a similarity measurement instead of using kernel-induced distance metric as in [28]. The XB validity metric used in this work was based on the modified objective function of the FCM algorithm, where the membership matrix does not need to be computed [34].

**PSO-based Dynamic Clustering:**
Omran et al. came up with an automatic hard clustering algorithm known as (DCPSO) [52-54]. The algorithm starts by partitioning the dataset into a relatively huge clusters number to decrease the influence of the initialization. By binary PSO, an optimal clusters number is chose. Various cluster validity indices such as Dunn index, S Dbw index and Turi index have been used as an objective function to measure the evolving process. Finally, the centreoids of the selection clusters are refined by the Kmeans technique. The authors applied the method for segmentation of multi-spectral, natural, synthetic images. They compared the performance of their proposed algorithm with unsupervised fuzzy approach (UFA) [46], the SOM approach [44], dynamic clustering using (DCGA), and dynamic clustering using random search (DCRS). They demonstrated that PSO-based dynamic clustering algorithm outperforms the others.

**HS-based Dynamic Clustering :**

Alia, Mandava [3] present a new dynamic clustering algorithm based on the hybridization of harmony search (HS) and fuzzy c-means to automatically segment MRI brain images in an intelligent manner. In this approach, the capability of standard HS is modified to automatically evolve the appropriate number of clusters as well as the locations of cluster centrees. By incorporating the concept of variable length encoding in each harmony memory vector, this algorithm is able to represent variable numbers of candidate cluster centrees at each iteration. Evaluation of the proposed algorithm has been performed using both real MRI data obtained from the Centree for Morphometric Analysis at Massachusetts General Hospital and simulated MRI data generated using the McGill University BrainWeb MRI simulator. Experimental results show the ability of this algorithm to find the appropriate number of naturally occurring regions in brain images. Furthermore, the superiority of the proposed algorithm over various state-of-the-art segmentation algorithms is demonstrated quantitatively.

## 4 Discussion

It is evident from the literature mentioned above that the metaheuristic -based algorithms are efficient in tackling the local optima problem of the partitional clustering. However, since these algorithms are heuristic and do not have any mathematical basis to support them when they are first introduced, their performance is proven through extensive experimentation. The challenge is in adapting them to the domain. Since their proof is through experimentation, researchers go a step further to continuously try several variations and combinations that may further improve the performance of these algorithms. This is clear from the GA-based clustering algorithms mentioned above where various modifications have been introduced. These modifications are even based on the optimization behavior, i.e., balancing between exploitation and exploration, of the algorithm by incorporating some local search mechanism or through using various evolving techniques such as encoding schemes (i.e., real or binary), selection mechanism or crossover and mutation operators. Furthermore, an investigation of the new metaheuristic algorithms to solve the optimization problems is also a target of the research committee even to prove its applicability or to improve solutions of these problems a bit further.

Based on what has been discussed earlier, there are three approaches that have been proposed in the literature to overcome the limitation of partitional clustering where the appropriate number of clusters in the given dataset is unknown a priori. However, metaheuristic based algorithms proved their capacity to solve such problem while the others have remained helpless as aforementioned. This is because metaheuristic algorithms are able to cope with local optima and explore large solution spaces with the ability to facilitate the prediction of the appropriate number of clusters in the given dataset. To cope with a different number of clusters, a modification to the solution representation in metaheuristic algorithm takes place with a utilization of cluster validity indices as an objective function. Despite the encouraging results obtained from the metaheuristic approach, it is evident from the literature mentioned above and from what has been reported in [35]that such approach is still in its early stage since the number of published articles is meager. Looking into these algorithms and into the foundations upon which they were based one can say that the competition in improving the performance of these algorithms for solving the clustering problem was the goal behind these developments. These developments were based on improving the natural behaviour of the optimization process of these algorithms. For instance, balancing between the exploration and exploitation strategies of metaheuristic algorithms is one way of improving the metaheuristic-based clustering algorithms. Also using various evolving techniques such as encoding schemes (i.e., real or binary), selection mechanism or crossover and mutation operators is another way of modifying these algorithms to improve their performance and accuracy. Furthermore, an investigation into the new metaheuristic algorithms to solve such problem is also another way to improve them. All of that is based on one fact about the metaheuristic approach that there is no available exact solution in such categories, therefore researchers go a step further to continuously try several variations and combinations that may further improve the performance of these algorithms.